\documentclass{article} 
\usepackage{iclr2023_conference,times}

\usepackage{hyperref}
\usepackage{url}

\usepackage{graphicx}
\usepackage{subfigure}
\usepackage{float}
\usepackage{amsmath}
\usepackage{amssymb}
\usepackage{multirow}
\usepackage{booktabs}
\usepackage{url}
\usepackage{array}
\usepackage{enumitem}
\usepackage{algorithm}
\usepackage{algorithmic}
\usepackage{pifont}
\usepackage{bm}
\usepackage{lipsum}
\usepackage{xcolor}
\usepackage{makecell}
\usepackage{CJKutf8}
\usepackage{listings}
\usepackage{bbm}

\hypersetup{
    colorlinks=true,
    linkcolor=blue,
    citecolor=blue,
    urlcolor=blue
}

\title{Rethinking LLM Language Adaptation: A Case Study on Chinese Mixtral}

\author{Yiming Cui \qquad Xin Yao \\ 
Joint Laboratory of HIT and iFLYTEK Research (HFL), Beijing, China \\
\texttt{ymcui@ieee.org\thanks{Email corresponding.}, yaoxin94@foxmail.com} \\
}

\iclrfinalcopy 
\begin{document}
\begin{CJK*}{UTF8}{gkai}

\maketitle

\begin{abstract}
Mixtral, a representative sparse mixture of experts (SMoE) language model, has received significant attention due to its unique model design and superior performance.
Based on Mixtral-8x7B-v0.1, in this paper, we propose Chinese-Mixtral and Chinese-Mixtral-Instruct with improved Chinese language abilities by adopting further pre-training and instruction fine-tuning.
Experimental results show that our Chinese-Mixtral and Chinese-Mixtral-Instruct successfully improve Chinese understanding and generation performance while retaining the original English abilities.
Then, we discuss several key questions when performing language adaptation on large language models, including the necessity of extending the language-specific vocabulary and the choice of the initialization model (foundation model v.s. instruction model), by providing empirical results and analysis.
We also present the visualizations of each expert to examine their importance on downstream tasks.
Our resources are publicly available through \url{https://github.com/ymcui/Chinese-Mixtral}.
\end{abstract}


\section{Introduction}

Natural Language Processing (NLP) has been revolutionized by introducing Large Language Models (LLMs) like the GPT series, which excel in generating human-like text.
These advancements, particularly with the development of ChatGPT \citep{chatgpt} and its successor, GPT-4 \citep{gpt-4}, have showcased not only remarkable achievements in language comprehension and generation but also in multi-modal and reasoning tasks, pushing the boundaries of what is possible in NLP and stirring interest in Artificial General Intelligence (AGI). 
Despite their transformative impact and the broadening of research and applications they have enabled, the proprietary nature and the significant computational resources required for these models pose challenges, limiting accessibility and hindering further innovation in the field by the broader research community.

To further promote open research, various open-source large language models have been proposed, and consequently related open-source community embraced significant advancement in recent days.
Among various open-source LLMs, LLaMA \citep{llama} and Llama-2 \citep{touvron2023llama2} have made unnegligible contributions to the research community, resulting in massive variants, such as Alpaca \citep{alpaca}, Chinese-LLaMA \citep{chinese-llama-alpaca}, etc.
Like GPT-2, LLaMA also adopts a decoder-only transformer architecture \citep{attention_is_all_you_need}, which has become a main-stream architecture for LLMs.

Mixtral \citep{jiang2024mixtral}, on the other hand, is a sparse mixture of experts (SMoE) model, which is different from LLaMA series.
Mixtral's architecture allows for efficient parameter utilization by employing a decoder-only setup where a router network selects two out of eight distinct groups of parameters (the ``experts'') for processing each token, combining their outputs additively. 
This approach not only increases model parameters efficiently but also optimizes cost and latency by using only a fraction of parameters per token, enabling faster inference at lower batch sizes and higher throughput at larger ones.
Mixtral exhibits strong performances on various benchmarks, surpassing Llama-2 70B and GPT-3.5, while Mixtral only activates 13B parameters at the inference stage.

Though LLaMA series and Mixtral model exhibit excellent performance in various benchmarks, they mainly focus on processing English, and consequently, their support for other languages is limited.
Following our previous attempt to adapt LLaMA and Llama-2 series into Chinese, in this paper, we propose Chinese-Mixtral and Chinese-Mixtral-Instruct, which are adapted from Mixtral-8x7B-v0.1, by performing continual training on Chinese text and instruction data.
To test the effectiveness of the proposed Chinese Mixtral models, we conduct various experiments, including automated benchmarks, human evaluations, etc.
We also provide several discussions and visualizations of the proposed model, which may shed light upon future research on creating effective fine-tuned LLMs.

The contributions of this paper can be summarized as follows.

\begin{itemize}[leftmargin=0.05\textwidth]
	\item We propose Chinese-Mixtral and Chinese-Mixtral-Instruct, which are further tuned on Mixtral-8x7B-v0.1 using QLoRA, and their effectiveness is validated by various experimental results, including automated benchmarks and human evaluation.
	\item We actively discuss several key questions by presenting point-to-point empirical experiments, including the necessity of extending vocabulary and the choice of the starting model, which may provide insights into creating better fine-tuned LLMs.
	\item We present various visualizations of each expert and discuss their roles in different downstream tasks, which may help better understand Mixtral.
	\item We have made our resources publicly available through GitHub, enabling open research and collaboration in our open-source community.
\end{itemize}

\section{Chinese Mixtral}

\begin{figure}[ht]
  \centering
  \includegraphics[width=0.8\columnwidth]{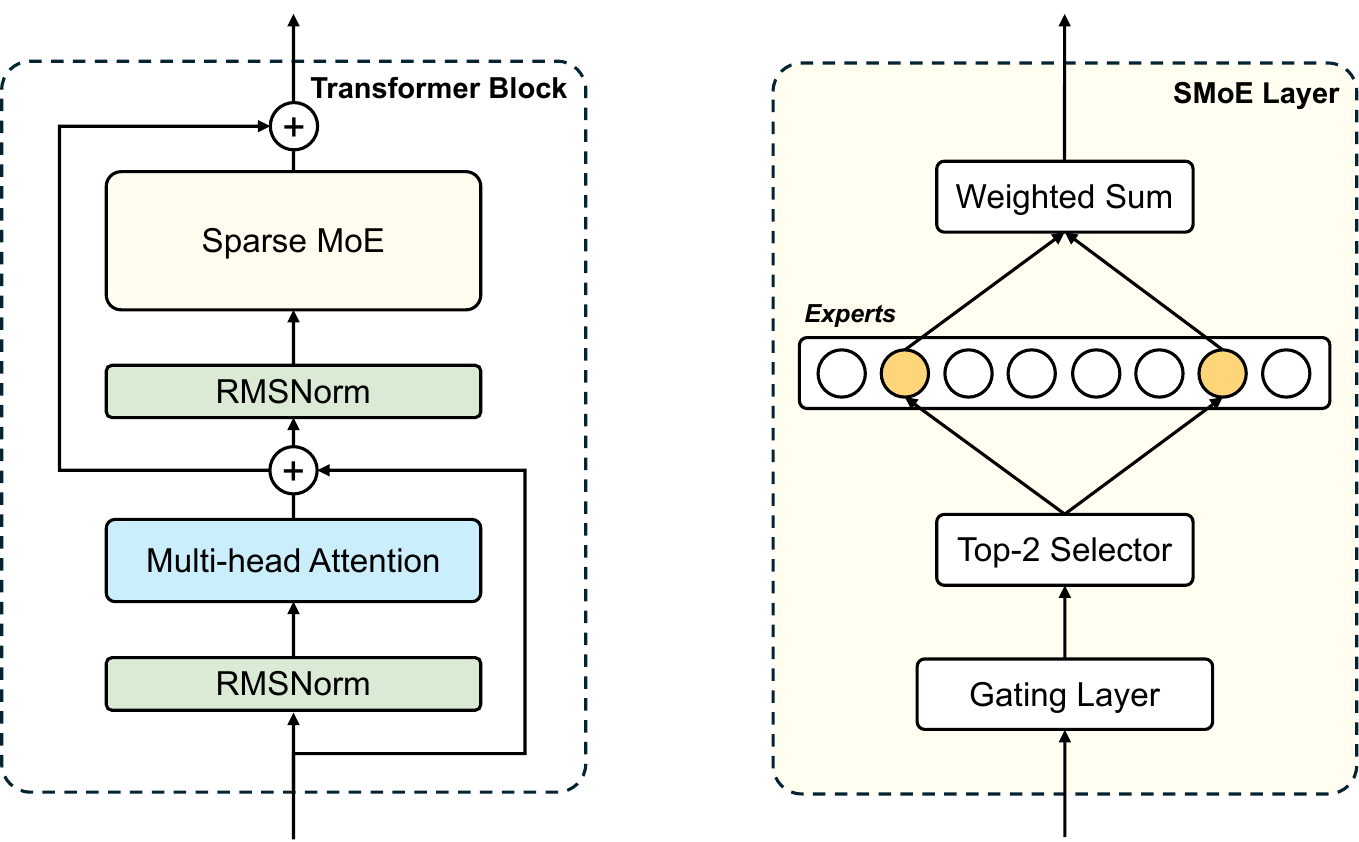}
  \caption{\label{mixtral-arch} Architecture of Mixtral. } 
\end{figure}

Our Chinese Mixtral models use exactly the same architecture as the original one without the vocabulary extension.
We briefly introduce the architecture and training method of Mixtral.
An overview of the Mixtral model is depicted in Figure \ref{mixtral-arch}.

Mixtral is built upon the transformer model \citep{attention_is_all_you_need}, where the feedforward layer is substituted by Sparse Mixture-of-Expert (SMoE).
The SMoE layer consists of 8 experts, and each of them is a feedforward layer.
A gating layer $G(x)$ is applied to select the most important two experts, as calculated by Equation \ref{eq-mixtral-gating}.
It is worth noting that the SMoE layer is applied in a token-level manner, where each token will have different combinations of experts.
\begin{equation}\label{eq-mixtral-gating}
	G(x) = \mathrm{Norm}(\mathrm{Top2}(\mathrm{Softmax} (x \cdot W_g))) , W_g \in \mathbb{R}^{d \times 8}
\end{equation}

The $\mathrm{Norm(\cdot)}$ is the normalization operation, where each element is divided by their sum.
Note that the Equation \ref{eq-mixtral-gating} is slightly different from the original paper's, where we mainly follow the actual implementation here.\footnote{\url{https://github.com/huggingface/transformers/blob/main/src/transformers/models/mixtral/modeling_mixtral.py}}
Finally, the output of the SMoE layer is obtained by the weighted sum of the corresponding Top-2 experts, where the SwiGLU architecture \citep{swiglu} is adopted in Mixtral for each expert.
\begin{equation}\label{eq-mixtral}
	y = \sum_{i=0}^{n-1} G(x)_i \cdot \mathrm{SwiGLU}_i(x)
\end{equation}

In the training phase, Mixtral applies additional auxiliary load balancing loss (as in \citet{fedus2022switch}) along with the loss of causal LM.
The auxiliary load balancing loss aims to penalize the cases where the routing between experts is extremely unbalanced.
Given a batch size of $B$ with $T$ tokens, the auxiliary loss is formulated as 
\begin{gather}\label{eq-aux-loss}
	\mathcal{L}_\text{aux} = 8 \cdot \sum_{i=0}^{n-1} f_i \cdot P_i , \\
	f_i = \frac{1}{T} \sum_{x \in B} \mathbbm{1} \{\arg\max p(x)=i\} ,
	P_i = \frac{1}{T} \sum_{x \in B} p_i(x),
\end{gather}
where $f_i$ is the fraction of tokens dispatched to expert $i$, and $P_i$ is the fraction of the router probability for expert $i$.
The final training loss is shown in Equation \ref{eq-training-loss}.
\begin{equation}\label{eq-training-loss}
	\mathcal{L} = \mathcal{L}_\text{clm} + \alpha \mathcal{L}_\text{aux}
\end{equation}

\section{Experimental Setups}

We largely follow the experimental setting as in our previous work \citep{chinese-llama-alpaca}.
Detailed settings are listed in Table \ref{experimental-settings}.
Chinese-Mixtral was initialized by Mixtral-8x7B-v0.1 and was further pre-trained using 20GB general Chinese corpus, which is identical to our Chinese-LLaMA base series.
Chinese-Mixtral-Instruct was built on Chinese-Mixtral and was further fine-tuned using 5M Chinese instruction data, which is identical to our Chinese-Alpaca-2 series.
Unlike our previous attempt, in this paper, {\bf we did not apply vocabulary extension to the original tokenizer}, i.e., we directly use the original Mixtral tokenizer. 
Further discussions can be found in Section \ref{sec-discussion-vocab-extend}.

All models are trained with the QLoRA method \citep{dettmers2024qlora}, where the embedding and LM head are fully trained (without using LoRA).
We employ the AdamW optimizer \citep{loshchilov2018decoupled} with a peak learning rate of 1e-4 with cosine scheduler and 5\% warm-up ratio. 
The scaling factor $\alpha$ of the auxiliary loss in Equation \ref{eq-training-loss} is set to 0.02 (default value in original Mixtral).
The implementation was done by using PEFT\footnote{\url{https://github.com/huggingface/peft}} and transformers\footnote{\url{https://github.com/huggingface/transformers}}.
We also utilize DeepSpeed \citep{deepspeed} to optimize memory efficiency during the training process. 
All models are trained on 48 A40 GPUs (48GB VRAM), where the pre-training takes one epoch and the SFT takes three epochs.
The total batch size is 1152.

\begin{table}[h]
\caption{\label{experimental-settings} {Hyperparameters and training details of Chinese-Mixtral and Chinese-Mixtral-Instruct. }}
\begin{center}
\begin{tabular}{l c c }
\toprule
\bf Settings & \bf Chinese-Mixtral & \bf Chinese-Mixtral-Instruct \\
\midrule
Trained From			& Mixtral-8x7B-v0.1	& Chinese-Mixtral \\
Training Data		& 20GB raw text ($\approx$7B tokens) & 5M Instruction SFT \\
Training Epochs		& 1 & 3 \\
Layer \# 			& \multicolumn{2}{c}{32} \\
Total Experts \# 	& \multicolumn{2}{c}{8} \\
Used Experts \#  	& \multicolumn{2}{c}{2} \\
Hidden Size 			& \multicolumn{2}{c}{4096} \\
Context Length 		& \multicolumn{2}{c}{32768} \\
Vocab Size			& \multicolumn{2}{c}{32000} \\
Training Method		& \multicolumn{2}{c}{QLoRA + Full training on embedding and LM head} \\
LoRA Rank			& \multicolumn{2}{c}{64} \\
LoRA Alpha			& \multicolumn{2}{c}{128} \\
LoRA Applied to		& \multicolumn{2}{c}{QKVO + W123 + Gate}  \\
\bottomrule
\end{tabular}
\end{center}
\end{table}

Though we use the context size of 1024 during the training phase, we are able to retain the original Mixtral context ability (i.e., 32K context length).
Moreover, by performing the quantitative analysis (Section \ref{sec-discuss-long-context}), we are surprised to see that the Mixtral model series (including our Chinese Mixtral models) are capable of handling way longer context length (up to 128K).

\section{Experimental Results}

In this section, we present experimental results on various automated benchmarks and chatbot arena (i.e., human evaluation).
Note that we mainly compare our results to Mixtral-8x7B-v0.1, from which our models were trained.
We also list Mixtral-8x7B-Instruct-v0.1 for comparison, though it is not used in our Chinese-Mixtral and Chinese-Mixtral-Instruct.

\subsection{C-Eval}

C-Eval \citep{huang2023ceval} is a multi-choice question answering dataset, which mainly covers four categories: STEM, Social, Humanities, and Others, consisting of nearly 14K samples for 52 disciplines.
Similar to other multi-choice QA datasets, such as RACE \citep{lai-etal-2017-race}, it requires the model to produce the correct option label based on the given question.
We tested our model on the validation split (1,346 samples) and test split (12,342 samples), where the test scores are obtained by submitting models' prediction files to the official leaderboard.
The results are shown in Table \ref{ceval-results}.
As we can see, the Mixtral series performs significantly better than other 13B models.
Chinese-Mixtral performs slightly worse than Mixtral-8x7B-v0.1.
However, after instruction fine-tuning, Chinese-Mixtral-Instruct brings significant improvements over Chinese-Mixtral, and even surpasses the original Mixtral-8x7B-Instruct-v0.1.

\begin{table}[h]
\caption{\label{ceval-results} {Results on C-Eval. }}
\begin{center}
\begin{tabular}{l c c c c }
\toprule
\multirow{2}*{\bf Model} & \multicolumn{2}{c}{\centering \bf Valid Set} & \multicolumn{2}{c}{\centering \bf Test Set} \\ 
 & \bf zero-shot & \bf 5-shot & \bf zero-shot & \bf 5-shot \\
\midrule
Chinese-LLaMA-2-13B		& 40.6 & 42.7 & 38.0 & 41.6 \\
Chinese-Alpaca-2-13B 	& 44.3 & 45.9 & 42.6 & 44.0 \\
Mixtral-8x7B-v0.1 		& 47.3 & 54.6 & 46.1 & 50.3 \\
Mixtral-8x7B-Instruct-v0.1 & 51.6 & 54.0 & 48.7 & 50.7 \\
\midrule
\bf Chinese-Mixtral 				& 45.8 & 54.2 & 43.1 & 49.1 \\
\bf Chinese-Mixtral-Instruct 	& 51.7 & 55.0 & 50.0 & 51.5 \\
\bottomrule
\end{tabular}
\end{center}
\end{table}

\subsection{CMMLU}

CMMLU \citep{li2023cmmlu} is another comprehensive Chinese evaluation dataset specifically designed to assess the knowledge and reasoning ability of language models in Chinese contexts.
CMMLU covers 67 topics, from basic subjects to advanced professional levels, with a total of 11.5K multiple-choice questions. 
The results are shown in Table \ref{cmmlu-results}.
Similar to the results in C-Eval, Chinese-Mixtral performs worse than the original Mixtral-8x7B-v0.1, and the Chinese-Mixtral-Instruct brings significant improvement, especially in the zero-shot settings (from 42.5 to 50.0).

\begin{table}[h]
\caption{\label{cmmlu-results} {Results on CMMLU. }}
\begin{center}
\begin{tabular}{l c c }
\toprule
\bf Model & \bf Test (zero-shot) & \bf Test (5-shot) \\
\midrule
Chinese-LLaMA-2-13B			& 38.9 & 42.5  \\
Chinese-Alpaca-2-13B 		& 43.2 & 45.5 \\
Mixtral-8x7B-v0.1 			& 44.3 & 51.6 \\
Mixtral-8x7B-Instruct-v0.1 	& 48.2 & 51.6 \\
\midrule
\bf Chinese-Mixtral				& 42.5 & 51.0 \\
\bf Chinese-Mixtral-Instruct 	& 50.0 & 53.0 \\
\bottomrule
\end{tabular}
\end{center}
\end{table}

\subsection{Open LLM Leaderboard}

We also submit our models on Hugging Face Open LLM Leaderboard\footnote{\url{https://huggingface.co/spaces/HuggingFaceH4/open_llm_leaderboard}}.
The leaderboard contains the following benchmarks (in English) to comprehensively test large language model abilities: ARC \citep{clark2018think}, HellaSwag \citep{zellers2019hellaswag}, MMLU \citep{hendrycks2020measuring}, TruthfulQA \citep{lin2021truthfulqa}, WinoGrande \citep{sakaguchi2021winogrande}, and GSM8K \citep{cobbe2021training}.
The results are shown in Table \ref{openllm-results}.

Overall, excluding GSM8K, our Chinese-Mixtral performs on par to the original Mixtral-8x7B-v0.1, which suggests that a large portion of English abilities are preserved.
On top of that, Chinese-Mixtral-Instruct further improves performance, especially for MMLU, TruthfulQA, and GSM8K.
We noticed a disastrous performance loss on GSM8K for Chinese-Mixtral, while we did not observe such phenomena on other tasks.
With an initial investigation, we discover that GSM8K requires to extract the answer after ``\#\#\#\#'' tokens, which is significantly different from other tasks.
In this context, our Chinese-Mixtral failed to follow the instruction, and the output text does not contain any ``\#\#\#\#'' tokens, consequently resulting in empty answers.
However, as we can see, Chinese-Mixtral-Instrct recovered GSM8K performance to a reasonable level.

This suggests that performing additional pre-training on LLM in other languages will result in unexpected performance loss on some downstream tasks, especially those requiring instruction-following abilities.
Such deficiency can be alleviated by adopting further instruction fine-tuning.

\begin{table}[h]
\caption{\label{openllm-results} {Results on Open LLM Leaderboard. HellaS: HellaSwag, TQA: TruthfulQA, WinoG: WinoGrande. }}
\small
\begin{center}
\begin{tabular}{l c c c c c c c}
\toprule
\bf Model & \bf ARC & \bf HellaS & \bf MMLU & \bf TQA & \bf WinoG & \bf GSM8K & \bf Average \\
\midrule
Chinese-LLaMA-2-13B 			& 55.80 & 79.53 & 53.01 & 38.24 & 75.69 & 3.94  & 51.04 \\
Chinese-Alpaca-2-13B 		& 58.70 & 79.76 & 55.12 & 50.22 & 75.61 & 25.02 & 57.41 \\
Mixtral-8x7B-v0.1			& 66.38 & 86.46 & 71.88 & 46.81 & 81.69 & 57.62 & 68.47 \\
Mixtral-8x7B-Instruct-v0.1 	& 70.14 & 87.55 & 71.40 & 64.98 & 81.06 & 61.11 & 72.70 \\
\midrule
\bf Chinese-Mixtral 				& 67.58 & 85.34 & 70.38 & 46.86 & 82.00 & 0.00 & 58.69 \\
\bf Chinese-Mixtral-Instruct 	& 67.75 & 85.67 & 71.53 & 57.46 & 83.11 & 55.65 & 70.19 \\
\bottomrule
\end{tabular}
\end{center}
\end{table}

\subsection{LongBench}

LongBench \citep{bai2023longbench} is a benchmark for evaluating the long-text understanding abilities of large language models.
LongBench consists of 6 categories and 20 different tasks, most of which have an average length of 5K-15K words, totaling about 4.75K test items.
We test our models on Chinese subsets and coding tasks.
The results are shown in Table \ref{longbench-results}.
As we can see, Chinese-Mixtral performs poorly compared to its initialization model, i.e., Mixtral-8x7B-v0.1, especially on summarization, code, and synthetic tasks.
However, after instruction fine-tuning, Chinese-Mixtral-Instruct performs significantly better than both models.
For example, in synthetic tasks, Chinese-Mixtral-Instruct scores 89.5, which is better than Chinese-Mixtral (14.0) and Mixtral-8x7B-v0.1 (83.5).
This demonstrates that though pre-training on different languages will temporarily harm the performance of downstream tasks, these issues will be significantly alleviated through proper instruction fine-tuning.
We also noticed that our Chinese-Mixtral-Instruct still leg behind the original Mixtral-8x7B-Instrct-v0.1, which is further trained on Mixtral-8x7B-v0.1 using instruction fine-tuning and direct preference optimization (DPO) \citep{rafailov2024direct}. 
In the future, we will try to also apply DPO on our Chinese-Mixtral-Instruct to see if it can receive additional performance gains.

\begin{table}[h]
\caption{\label{longbench-results} {Results on LongBench (Chinese + code tasks). S-QA: Single-doc QA, M-QA: Multi-doc QA, Summ: Summarization, FS-Learn: Few-shot Learning, Code: Code Completion, Synthetic: Synthetic Tasks.}}
\small
\begin{center}
\begin{tabular}{l c c c c c c c}
\toprule
\bf Model & \bf S-QA & \bf M-QA & \bf Summ & \bf FS-Learn & \bf Code & \bf Synthetic & \bf Average \\
\midrule
Chinese-LLaMA-2-7B-64K 		& 27.2 & 16.4 & 6.5  & 33.0 & 7.8  & 5.0  & 16.0 \\
Chinese-Alpaca-2-7B-64K 		& 44.7 & 28.1 & 14.4 & 39.0 & 44.6 & 5.0  & 29.3 \\
Chinese-LLaMA-2-13B-16K		& 36.7 & 17.7 & 3.1  & 29.8 & 13.8 & 3.0  & 17.3 \\
Chinese-Alpaca-2-13B-16K		& 47.9 & 26.7 & 13.0 & 22.3 & 46.6 & 21.5 & 29.7 \\
Mixtral-8x7B-v0.1 			& 35.5 & 9.5  & 16.4 & 46.5 & 57.2 & 83.5 & 41.4 \\
Mixtral-8x7B-Instruct-v0.1  & 56.5 & 35.7 & 15.4 & 46.0 & 63.6 & 98.0 & 52.5 \\
\midrule
\bf Chinese-Mixtral 				& 32.0 & 23.7 & 0.4  & 42.5 & 27.4 & 14.0 & 23.3 \\
\bf Chinese-Mixtral-Instruct 	& 50.3 & 34.2 & 16.4 & 42.0 & 56.1 & 89.5 & 48.1 \\
\bottomrule
\end{tabular}
\end{center}
\end{table}

\subsection{Chinese LLM Chatbot Arena}

\begin{figure}[ht]
  \centering
  \includegraphics[width=1\columnwidth]{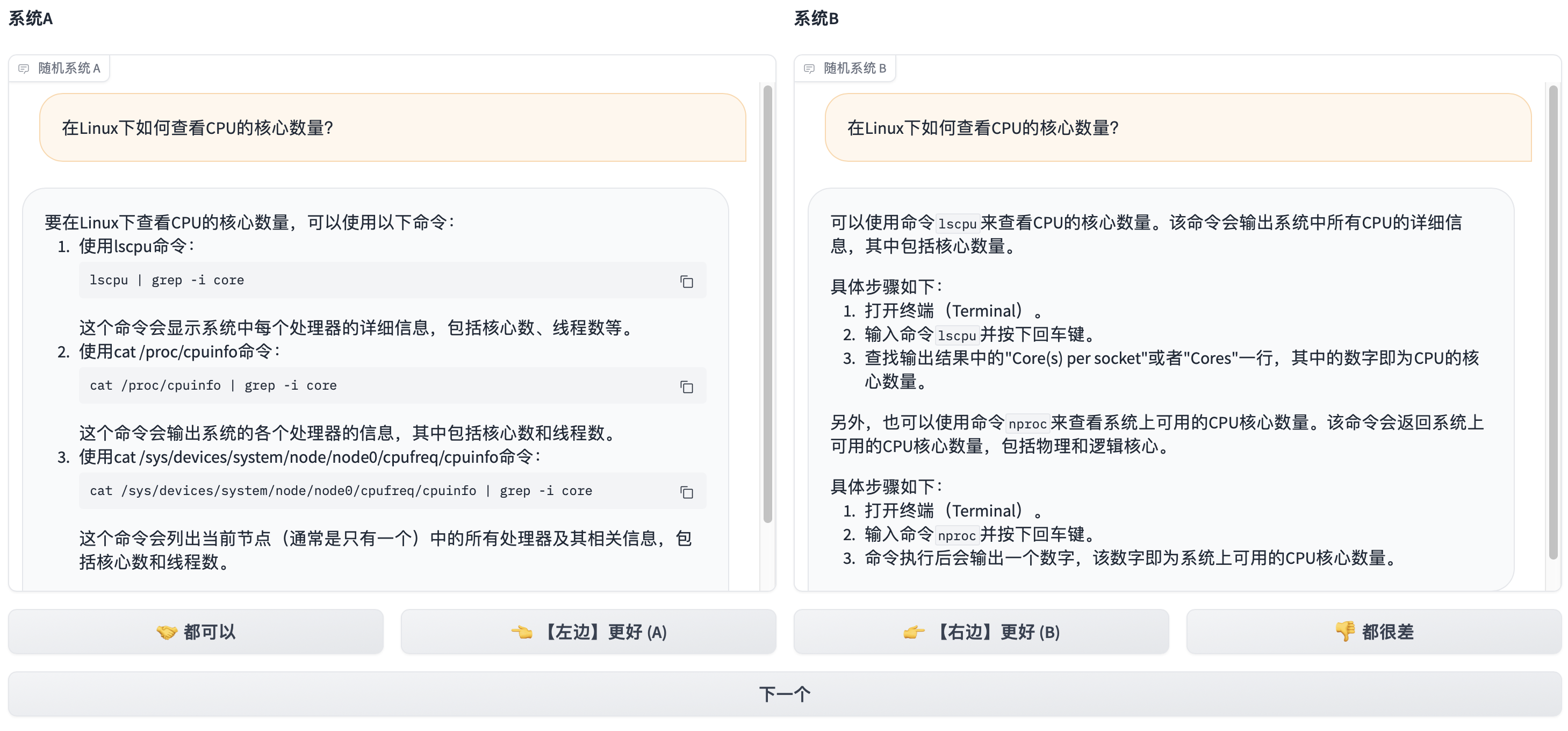}
  \caption{\label{arena-interface} Interface of Chinese LLM chatbot arena. } 
\end{figure}

\begin{figure*}[ht]
  \centering
  \subfigure[Winning rate (excluding ties)]{\includegraphics[width=0.48\textwidth]{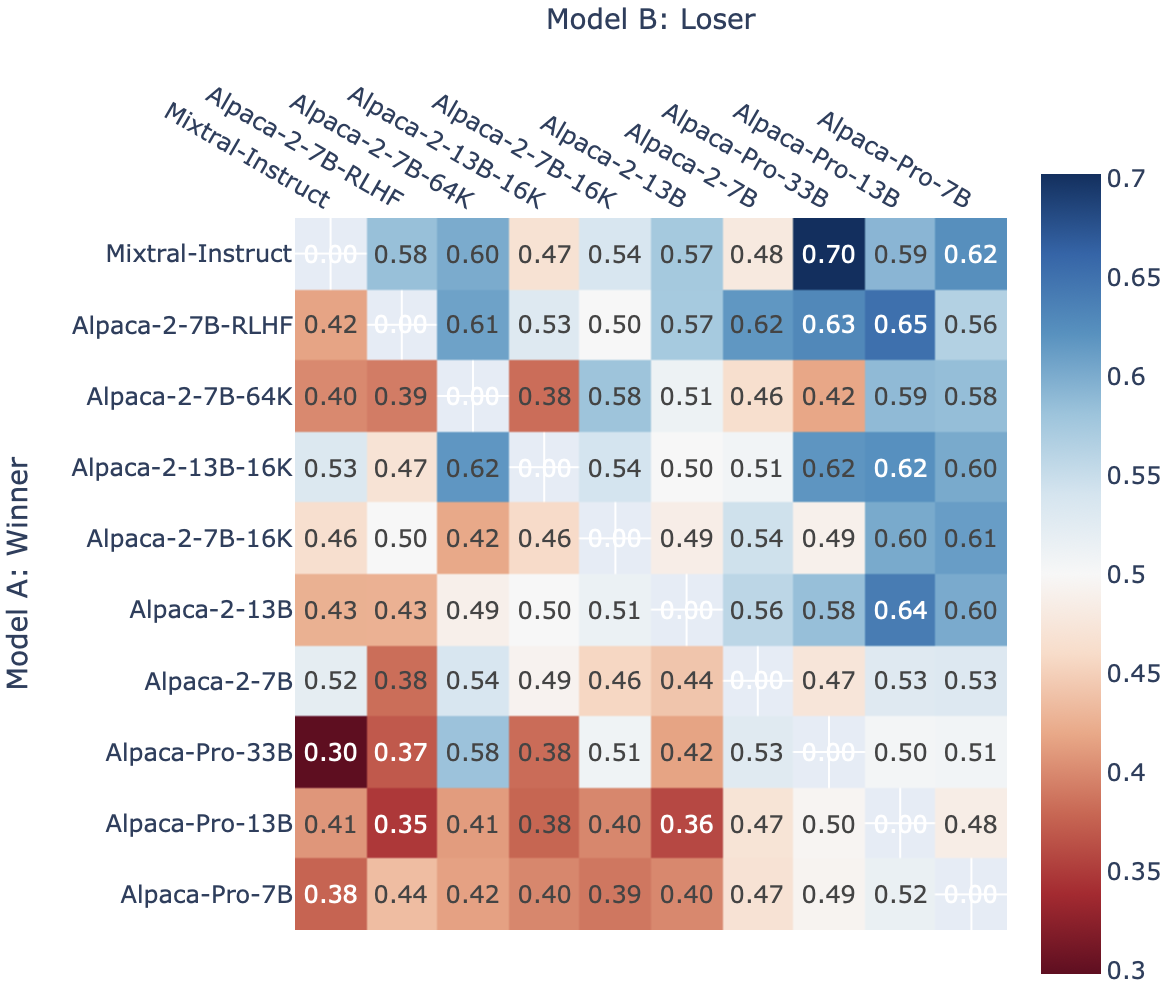}} 
  \subfigure[Battle count]{\includegraphics[width=0.48\textwidth]{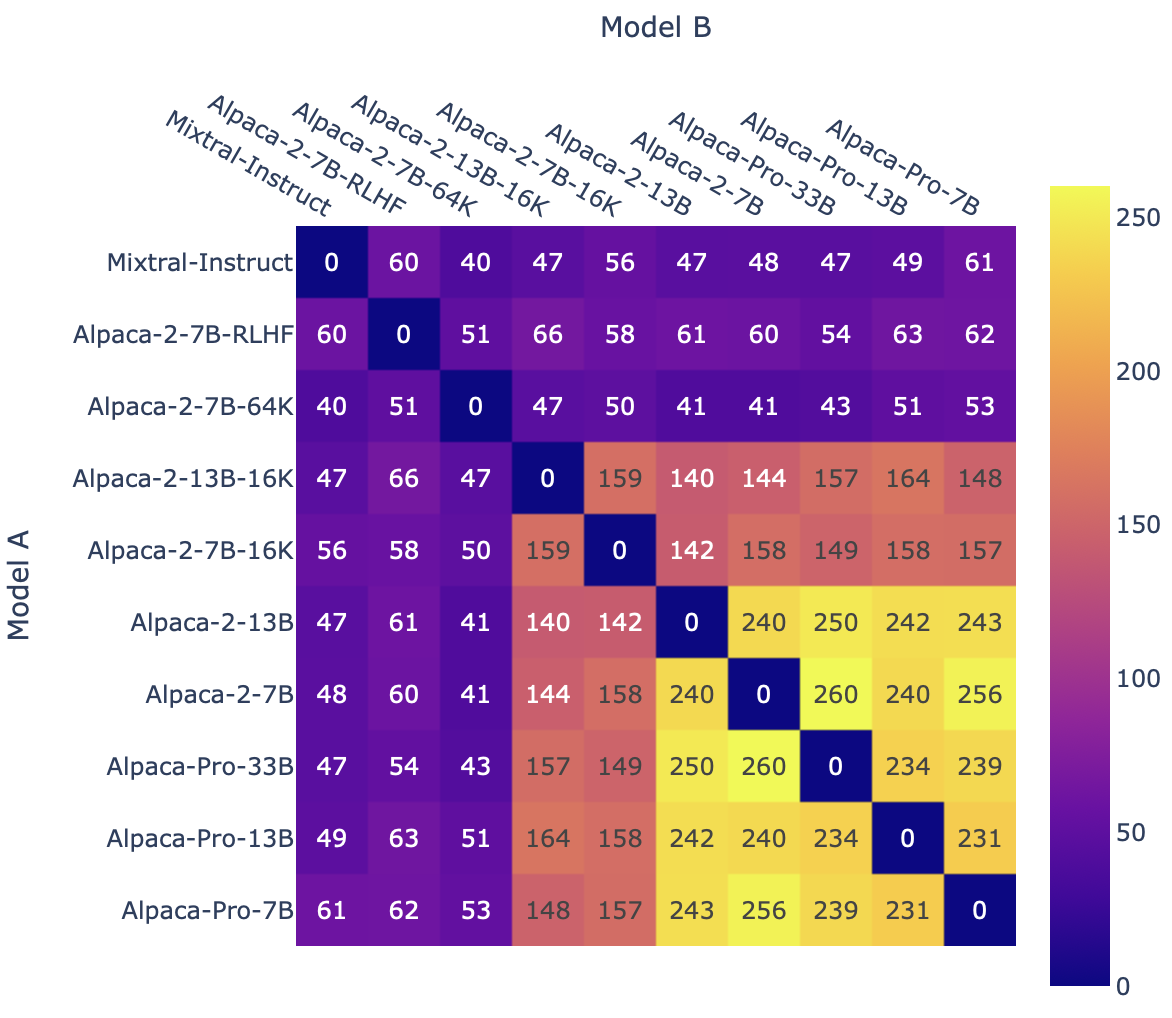}}
  \caption{\label{fig-chatbot-arena} Results of pair-wise winning rate and battle count in Chinese LLM chatbot arena. } 
\end{figure*}

In order to test the generation quality, we set up an online chatbot arena\footnote{\url{http://llm-arena.ymcui.com}} based on Gradio \citep{Abid_Gradio_Hassle-free_sharing_2019}, which contains the outputs of different LLMs w.r.t various instructions.
The instruction set contains 360 samples, covering different topics and tasks, such as question answering, math and reasoning, culture, translation, ethics, etc.
Each LLM retains three outputs for each instruction to preserve the diversity and reduce the randomness.
During human rating, two random anonymous LLMs will be shown w.r.t. the same instruction, and the user should rate which one is better (or equally better/worse). 
After rating, the user will be notified of the identity of the LLMs.
The user interface is shown in Figure \ref{arena-interface}, and the results are shown in Table \ref{arena-results}.
There are two metrics provided: 1) win rate (excluding ties); 2) Elo rating (the initial rating is 1500).
As we can see Chinese-Mixtral-Instruct yields the best performance on both metrics.
We also provide the pair-wise winning rate and the battle count in Figure \ref{fig-chatbot-arena}.

\begin{table}[h]
\caption{\label{arena-results} {Results of Chinese LLM chatbot arena (as of Feb 27, 2024). }}
\begin{center}
\begin{tabular}{l c c }
\toprule
\bf Model & \bf Win Rate (no ties) $\downarrow$ & \bf Elo Rating \\
\midrule
Chinese-Mixtral-Instruct		& 57.36\% & 1571 \\
Chinese-Alpaca-2-7B-RLHF 	& 56.45\% & 1474 \\
Chinese-Alpaca-2-13B			& 56.19\% & 1451 \\
Chinese-Alpaca-2-13B-16K		& 56.16\% & 1496 \\
Chinese-Alpaca-2-7B-16K		& 52.25\% & 1563 \\
Chinese-Alpaca-2-7B			& 48.72\% & 1497 \\
Chinese-Alpaca-2-7B-64K		& 48.44\% & 1536 \\
Chinese-Alpaca-Pro-33B		& 47.10\% & 1515 \\
Chinese-Alpaca-Pro-7B		& 44.55\% & 1437 \\
Chinese-Alpaca-Pro-13B		& 43.02\% & 1455 \\
\bottomrule
\end{tabular}
\end{center}
\end{table}

\section{Discussion}

\subsection{Effect of Extending Chinese Vocabulary}\label{sec-discussion-vocab-extend}

Extending vocabulary with language-specific entries has become a traditional routine of adapting English-based LLMs to other languages.
One of the main advantages is that vocabulary extension could significantly improve encoding efficiency in the target language.
For example, when adapting LLaMA/Mixtral to Chinese, the original vocabulary contains only very few Chinese characters as shown in Table \ref{vocab-efficiency-results}, where the Chinese characters only take up 2.2\% and 4.6\% respectively\footnote{We roughly identify Chinese characters using the following Unicode range: 
1) basic CJK unified ideographs ({\tt 4E00-9FFF}); 
2) CJK unified ideographs extension A ({\tt 3400-4DBF});
3) CJK unified ideographs extension B ({\tt 20000-2A6DF}).}.

We follow our previous attempt \citep{chinese-llama-alpaca} to the Mixtral model to add additional Chinese tokens into the original Mixtral tokenizer. 
Similarly, we use the pre-training data to train a Chinese tokenizer and then merge it with the original Mixtral vocabulary by removing the duplicate entries.
As we can see from Table \ref{vocab-efficiency-results}, the proportion of Chinese tokens in the vocabulary grows significantly, from 4.6\% to 44.6\%.

To test the encoding efficiencies, we use different tokenizers to tokenize the Chinese portion of Wikipedia data (wikipedia-zh-20231101)\footnote{\url{https://huggingface.co/datasets/wikimedia/wikipedia/tree/main/20231101.zh}}, where the dataset size is about 1.7GB.
We can see that after using Chinese tokenizers (Chinese-LLaMA-2 and Chinese-Mixtral-ext), the number of encoded tokens drops significantly, resulting in a 43.9\% and 32.6\% decline respectively, which means that the encoding efficiency significantly improved.

\begin{table}[h]
\caption{\label{vocab-efficiency-results} {Encoding results of Wikipedia-zh-20231101 using different tokenizers. ``-ext'' means the Chinese vocabulary extended tokenizer. }} 
\begin{center}
\begin{tabular}{l c c}
\toprule
\bf Tokenizer 	& \bf All / Zh Vocab \# & \bf Encoded Tokens \# \\
\midrule
Llama-2 				& 32,000 / 700 (2.2\%) & 1,432,314,226 \\	
Chinese-LLaMA-2			& 55,296 / 23,933 (43.3\%) & ~~~~~~~~~~~~~~803,138,206 \tiny{\textcolor{blue}{(-43.9\%$\downarrow$)}} \\
\midrule
Chinese-Mixtral 			& 32,000 / 1,459 (4.6\%) & 1,179,694,379 \\
Chinese-Mixtral-ext		& ~~55,296 / 24,644 (44.6\%) & ~~~~~~~~~~~~~~795,217,098  \tiny{\textcolor{blue}{(-32.6\%$\downarrow$)}}\\
\bottomrule
\end{tabular}
\end{center}
\end{table}

However, leaving encoding efficiency aside, {\bf \em does vocabulary extension also improve model performance?} 
In this paper, we provide a point-to-point comparison on this matter, which was not well-studied in the previous literature.
Using the extended Mixtral tokenizer, we use exactly the same training recipe to train Chinese-Mixtral-ext (foundation model) and Chinese-Mixtral-Instruct-ext (instruction model).
The embeddings of the newly added tokens are initialized with the average of their subtokens (using the original tokenizer), which we found to be more effective than random initialization.
After obtaining these models, we test their performances on C-Eval, CMMLU, and MMLU datasets to see if the extended vocabulary could improve performance on downstream tasks.
The results are shown in Table \ref{extend-vocab-results}.

\begin{table}[h]
\caption{\label{extend-vocab-results} {Results of Chinese vocabulary extension. 5-shot results are reported for all experiments. Differences to the counterparts are shown in parentheses. }}
\begin{center}
\begin{tabular}{l c c c c}
\toprule
\bf Model 	& \bf Vocab Size & \bf C-Eval & \bf CMMLU & \bf MMLU \\
\midrule
Chinese-Mixtral 				& 32,000 & 54.2 & 51.0 & 67.1 \\
Chinese-Mixtral-Instruct		& 32,000 & 55.0 & 53.0 & 69.6 \\	
\midrule
Chinese-Mixtral-ext			& 55,296 & 48.9 \tiny{\textcolor{blue}{(-5.3$\downarrow$)}} & 46.5 \tiny{\textcolor{blue}{(-4.5$\downarrow$)}}& 65.8 \tiny{\textcolor{blue}{(-1.3$\downarrow$)}} \\
Chinese-Mixtral-Instruct-ext& 55,296 & 52.5 \tiny{\textcolor{blue}{(-2.5$\downarrow$)}} & 51.7 \tiny{\textcolor{blue}{(-1.3$\downarrow$)}} & 68.6 \tiny{\textcolor{blue}{(-1.0$\downarrow$)}} \\
\bottomrule
\end{tabular}
\end{center}
\end{table}

Unfortunately, vocabulary extension does not bring performance improvements on both Chinese and English tasks.
For example, vocabulary-extended Chinese-Mixtral (i.e., Chinese-Mixtral-ext) results in a significant drop in Chinese tasks, where 5.3 (48.9 v.s. 54.2) and 4.5 (46.5 v.s. 51.0) performance drops can be seen for C-Eval and CMMLU, respectively.
Compared to Chinese tasks, we also observe a noticeable performance drop in MMLU, though the gap is not as big as in Chinese tasks.
After applying further instruction fine-tuning, Chinese-Mixtral-Instruct-ext obtains significant improvements in all tasks.
However, compared to Chinese-Mixtral-Instruct, each task still has a performance gap.

These experimental results reveal that vocabulary extension might not be a necessity when adapting English-based LLMs to other languages.
Though it can accelerate the encoding efficiency, it might not bring advantages in downstream tasks compared to the one that uses the original tokenizer.

\subsection{Effect of the Initialization Model}

When performing language ability transfer or further fine-tuning, we often encounter one key question: {\bf \em should I initialize with the foundation model or instruction model?}

For example, in terms of Mixtral model, we can either choose to train from Mixtral-8x7B-v0.1 or Mixtral-8x7B-Instruct-v0.1.
To investigate the question above, we conduct additional empirical experiments on Mixtral-8x7B-Instruct-v0.1 using exactly the same training setting as in our Chinese-Mixtral and Chinese-Mixtral-Instruct, which are trained on Mixtral-8x7B-v0.1.
The results are shown in Table \ref{starting-model-results}.
As we can see that,
\begin{itemize}
	\item Mixtral-8x7B-Instruct-v0.1 performs significantly better than Mixtral-8x7B-v0.1 in MMLU but not in C-Eval and CMMLU, which demonstrates that supervised fine-tuning on English instruction data can bring additional improvements on English downstream tasks, but it will add little benefits to Chinese tasks.
	\item After training with Chinese text on Mixtral-8x7B-Instruct-v0.1, we noticed significant drops in all downstream tasks compared to those experiments on Mixtral-8x7B-v0.1. By continuing instruction fine-tuning, the performances of downstream tasks improved, but the overall performances are still behind Chinese-Mixtral-Instruct.
\end{itemize}

The above observations reveal that it is preferred to start with the foundation model (Mixtral-8x7B-v0.1) rather than the instruction model (Mixtral-8x7B-Instruct-v0.1) when performing language ability transfer. 

\begin{table}[h]
\caption{\label{starting-model-results} {Results of using different starting model. 5-shot results are reported for all experiments. Differences to the counterparts are shown in parentheses. }}
\begin{center}
\begin{tabular}{l c c c}
\toprule
\bf Model & \bf C-Eval & \bf CMMLU & \bf MMLU \\
\midrule
\em Mixtral-8x7B-v0.1 				& 54.6 & 51.6 & 69.0 \\
+ pre-training (Chinese-Mixtral) 	& 54.2 & 51.0 & 67.1 \\	
++ SFT (Chinese-Mixtral-Instruct)	& 55.0 & 53.0 & 69.6 \\
\midrule
\em Mixtral-8x7B-Instruct-v0.1 		& 54.0 \tiny{\textcolor{blue}{(-0.6$\downarrow$)}} & 51.6 \tiny{(0.0)} & 70.4 \tiny{\textcolor{red}{(+1.4$\uparrow$)}} \\
+ pre-training 						& 52.5 \tiny{\textcolor{blue}{(-1.7$\downarrow$)}}& 50.0 \tiny{\textcolor{blue}{(-1.0$\downarrow$)}} & 66.1 \tiny{\textcolor{blue}{(-1.0$\downarrow$)}} \\
++ SFT 								& 52.7 \tiny{\textcolor{blue}{(-2.3$\downarrow$)}} & 51.7 \tiny{\textcolor{blue}{(-1.3$\downarrow$)}} & 67.6 \tiny{\textcolor{blue}{(-2.0$\downarrow$)}} \\
\bottomrule
\end{tabular}
\end{center}
\end{table}

\subsection{Effect of Long Context Abilities}\label{sec-discuss-long-context}

\begin{figure}[ht]
  \centering
  \includegraphics[width=0.95\columnwidth]{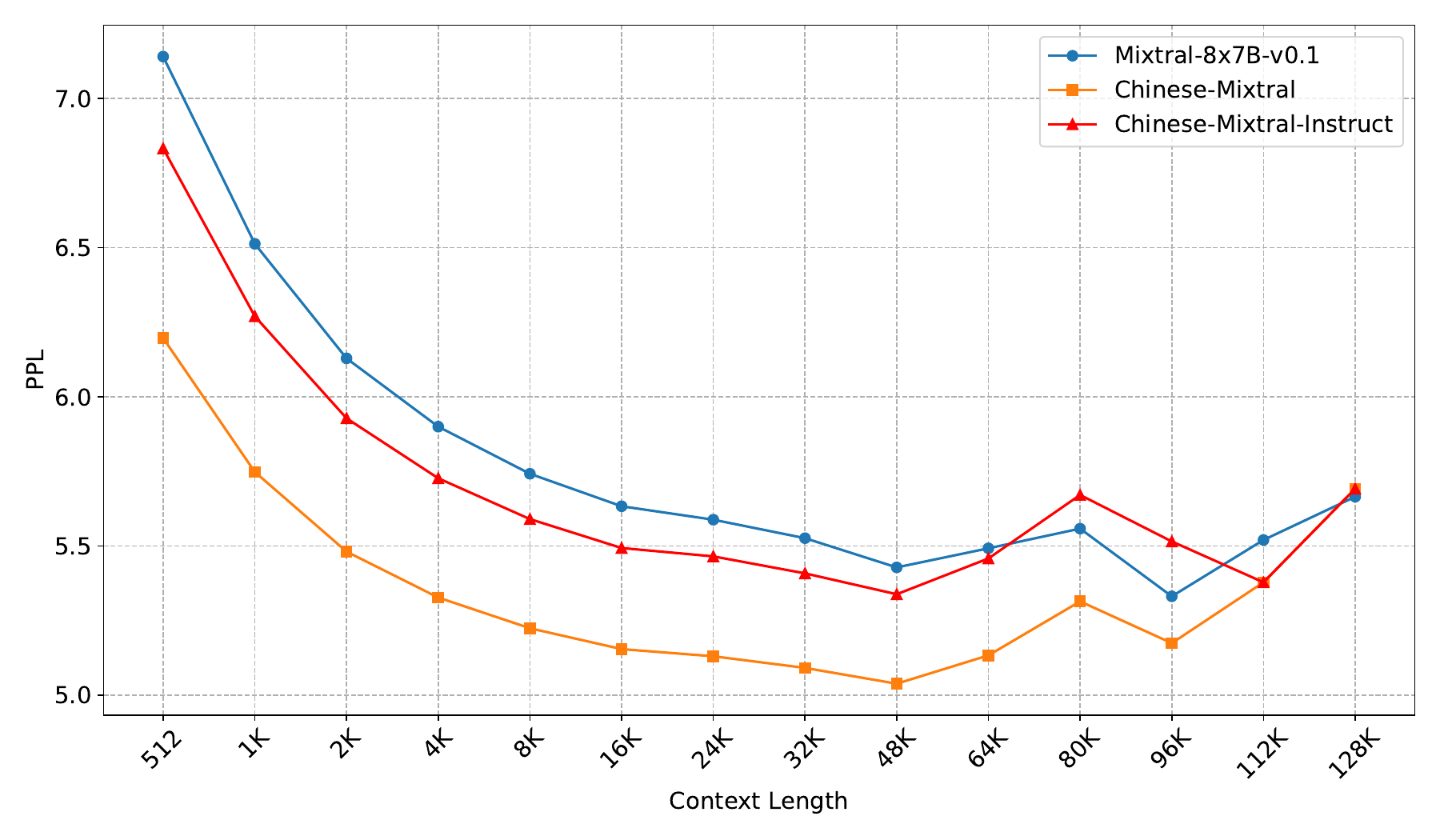}
  \caption{\label{ppl-comparison} Perplexities under different context lengths (on validation set). } 
\end{figure}

The original paper of Mixtral \citep{jiang2024mixtral} reports that the Mixtral supports a context size of 32,768 (32K). 
However, we wonder whether it can support longer context beyond 32K.
We plot the PPL under different context lengths on the pre-training validation set. 
As we did not perform vocabulary extension, the perplexities of Mixtral-8x7B-v0.1, Chinese-Mixtral, and Chinese-Mixtral-Instruct are directly comparable.
Although PPL can not be regarded as a comprehensive metric to evaluate LLMs, it is a general starting point to test LLM's basic abilities.
The results are shown in Figure \ref{ppl-comparison}.
As we can see, the perplexities continually go down when context lengths grow, and the optimal PPLs are in 48K but not 32K.
We are also surprised to see that these models still exhibit decent PPLs beyond 48K (even in 128K). 
This demonstrates that Mixtral series has a good long-context generalization ability, which may not require additional long-context tuning (such as PI \citep{chen2023extending} and YaRN \citep{peng2023yarn}, etc.) to support longer context.

\section{Visualizations}

\begin{figure*}[ht]
  \centering
  \subfigure[C-Eval performance]{\label{fig-visualization-a}\includegraphics[width=0.24\textwidth]{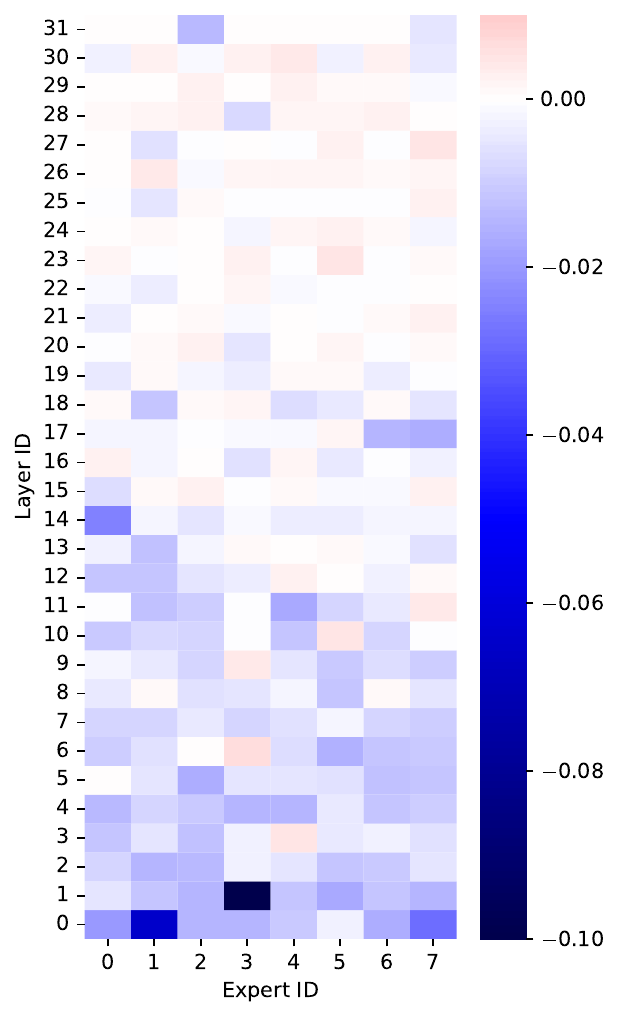}} 
  \subfigure[PPL performance]{\label{fig-visualization-b}\includegraphics[width=0.24\textwidth]{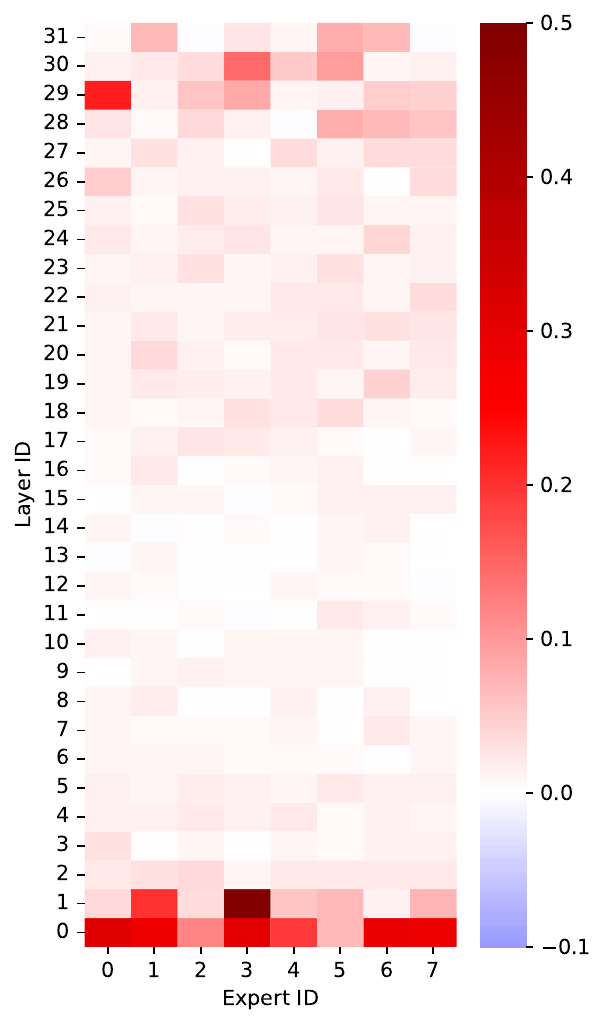}} 
  \subfigure[C-Eval throughput]{\label{fig-visualization-c}\includegraphics[width=0.24\textwidth]{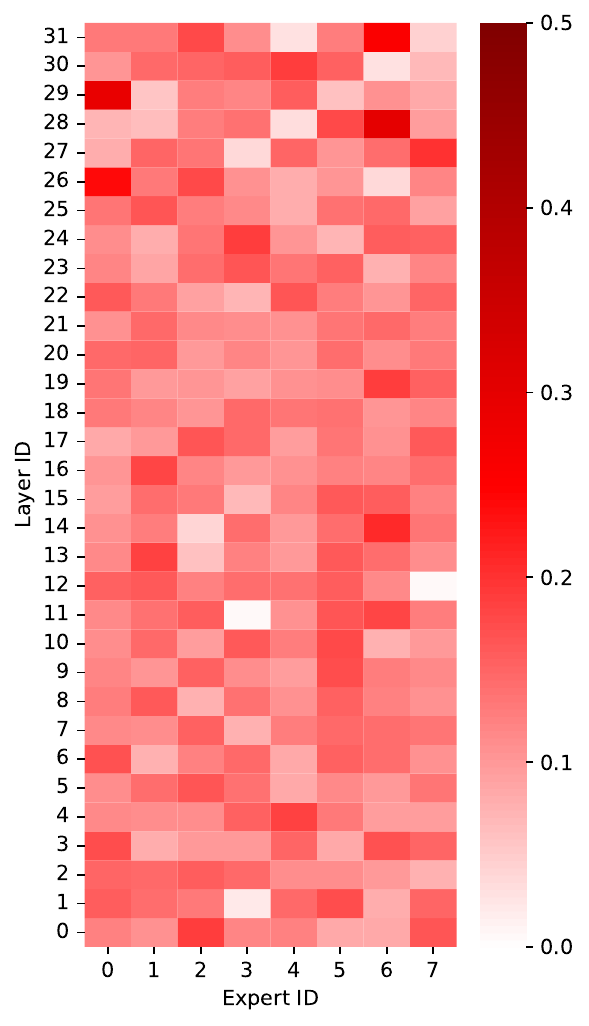}} 
  \subfigure[PPL throughput]{\label{fig-visualization-d}\includegraphics[width=0.24\textwidth]{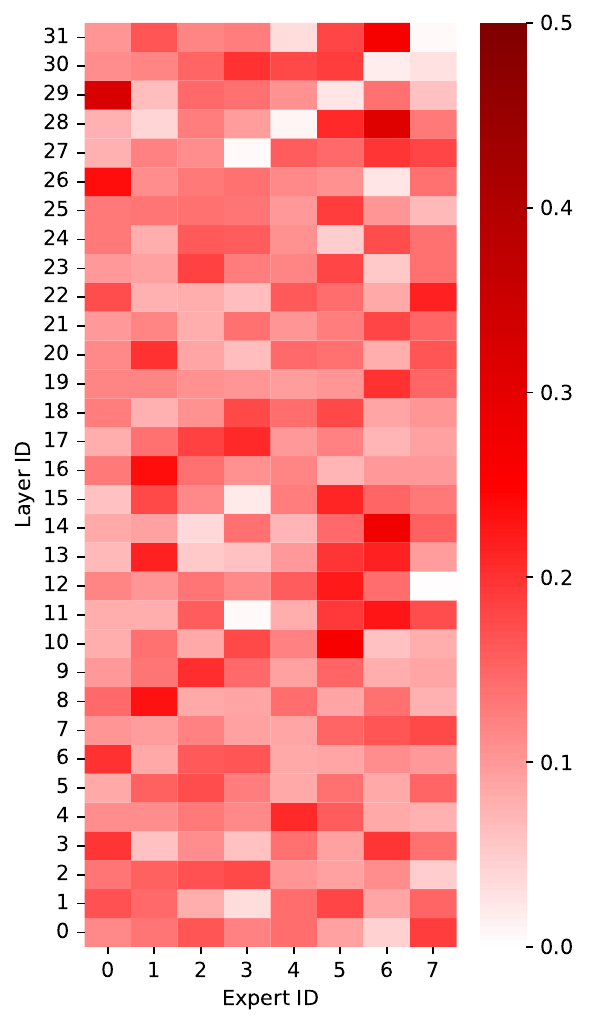}} \\  
  \caption{\label{fig-visualization} Visualizations of each expert in each layer.}
\end{figure*}

In this section, we analyze the importance of each expert in each layer by performing visualizations on Chinese-Mixtral.\footnote{We also conduct visualizations on Chinese-Mixtral-Instruct, and the resulting figures are similar.
We only show the visualizations of Chinese-Mixtral here for simplicity.}
Following the visualization method in \cite{cui-etal-2022-iscience}, we first disable a single expert (i.e., never be chosen by the router) and perform inference on downstream tasks.
The performance difference to the Chinese-Mixtral is visualized. 
That is
\begin{quote}
\begin{em}
cell\_value	= performance of disabling current expert - baseline performance.
\end{em}
\end{quote}
In this way, we can understand whether a particular expert is important to the overall performance.
We mainly visualize C-Eval performance (Figure \ref{fig-visualization-a}) and PPL on a small held-out set (Figure \ref{fig-visualization-b}). 
We have the following observations.
\begin{itemize}
	\item In Figure \ref{fig-visualization-a}, removing experts in lower layers has a larger impact (shown in blue, as they are lower than the baseline performance). In the upper layers, disabling some experts results in a better performance (shown in red), such as expert 7 in layer 27, etc. 
	\item In Figure \ref{fig-visualization-b}, removing experts in lower layers also has a significant impact, similar to the ones in C-Eval. However, we also observe that some of the experts in the upper layer also have great impacts, such as expert 0 in layer 29. The experts in the middle layers are not as important as the others. Also, in contrast to the results of C-Eval, we did not observe disabling any expert will bring a positive effect (as there is no cell marked in blue, i.e., lower PPL).
\end{itemize}

This observation suggests that the experts in the lower layers are the most important ones. 
The importance of experts in the middle and upper layers varies in different tasks.

By comparing Figure \ref{fig-visualization-a} and \ref{fig-visualization-b}, we notice that expert 3 in layer 1 is the most important one in both tasks.
To further investigate the most important expert (expert 3 in layer 1), we also plot the token throughput rate in Figure \ref{fig-visualization-c} and \ref{fig-visualization-d} for C-Eval and PPL test, respectively.
The token throughput rate is measured by calculating the total token number of passing a specific expert, and these numbers are then normalized in each layer.
To our surprise, we find that expert 3 is not the expert that processes the most tokens in layer 1 (as shown in lighter colors than the others in the same layer), which is identical in both tasks.
This observation suggests that the processed token number might not be a direct indicator of measuring the expert importance.
 
\section{Conclusion}

In this paper, we propose two Chinese-adapted Mixtral models, namely Chinese-Mixtral and Chinese-Mixtral-Instruct.
Different from our previous attempt on Chinese-LLaMA, we directly perform Chinese pre-training and instruction fine-tuning on Mixtral (Mixtral-8x7B-v0.1) without extending the vocabulary with additional Chinese tokens.
The experimental results show that Chinese-Mixtral and Chinese-Mixtral bring significant improvements in Chinese downstream tasks over the original Mixtral, and some of the results are even better than the original Mixtral-Instruct.
We discover that though extending Chinese vocabulary may bring notable improvements on encoding efficiency but it may not necessarily bring performance improvements on downstream tasks.
We also examine the effect of the initialization model and long context abilities, which might be useful in future work.
We have open-sourced our models and code to further facilitate open research and collaborations in our research community.

\bibliography{iclr2023_conference}

\begin{thebibliography}{28}
\providecommand{\natexlab}[1]{#1}
\providecommand{\url}[1]{\texttt{#1}}
\expandafter\ifx\csname urlstyle\endcsname\relax
  \providecommand{\doi}[1]{doi: #1}\else
  \providecommand{\doi}{doi: \begingroup \urlstyle{rm}\Url}\fi

\bibitem[Abid et~al.(2019)Abid, Abdalla, Abid, Khan, Alfozan, and Zou]{Abid_Gradio_Hassle-free_sharing_2019}
Abubakar Abid, Ali Abdalla, Ali Abid, Dawood Khan, Abdulrahman Alfozan, and James Zou.
\newblock {Gradio: Hassle-free sharing and testing of ML models in the wild}, 2019.
\newblock URL \url{https://arxiv.org/abs/1906.02569}.

\bibitem[Bai et~al.(2023)Bai, Lv, Zhang, Lyu, Tang, Huang, Du, Liu, Zeng, Hou, Dong, Tang, and Li]{bai2023longbench}
Yushi Bai, Xin Lv, Jiajie Zhang, Hongchang Lyu, Jiankai Tang, Zhidian Huang, Zhengxiao Du, Xiao Liu, Aohan Zeng, Lei Hou, Yuxiao Dong, Jie Tang, and Juanzi Li.
\newblock Longbench: A bilingual, multitask benchmark for long context understanding.
\newblock \emph{ArXiv preprint}, abs/2308.14508, 2023.
\newblock URL \url{https://arxiv.org/abs/2308.14508}.

\bibitem[Chen et~al.(2023)Chen, Wong, Chen, and Tian]{chen2023extending}
Shouyuan Chen, Sherman Wong, Liangjian Chen, and Yuandong Tian.
\newblock Extending context window of large language models via positional interpolation.
\newblock \emph{ArXiv preprint}, abs/2306.15595, 2023.
\newblock URL \url{https://arxiv.org/abs/2306.15595}.

\bibitem[Clark et~al.(2018)Clark, Cowhey, Etzioni, Khot, Sabharwal, Schoenick, and Tafjord]{clark2018think}
Peter Clark, Isaac Cowhey, Oren Etzioni, Tushar Khot, Ashish Sabharwal, Carissa Schoenick, and Oyvind Tafjord.
\newblock Think you have solved question answering? try arc, the ai2 reasoning challenge.
\newblock \emph{ArXiv preprint}, abs/1803.05457, 2018.
\newblock URL \url{https://arxiv.org/abs/1803.05457}.

\bibitem[Cobbe et~al.(2021)Cobbe, Kosaraju, Bavarian, Chen, Jun, Kaiser, Plappert, Tworek, Hilton, Nakano, et~al.]{cobbe2021training}
Karl Cobbe, Vineet Kosaraju, Mohammad Bavarian, Mark Chen, Heewoo Jun, Lukasz Kaiser, Matthias Plappert, Jerry Tworek, Jacob Hilton, Reiichiro Nakano, et~al.
\newblock Training verifiers to solve math word problems.
\newblock \emph{ArXiv preprint}, abs/2110.14168, 2021.
\newblock URL \url{https://arxiv.org/abs/2110.14168}.

\bibitem[Cui et~al.(2022)Cui, Zhang, Che, Liu, Chen, and Wang]{cui-etal-2022-iscience}
Yiming Cui, Wei-Nan Zhang, Wanxiang Che, Ting Liu, Zhigang Chen, and Shijin Wang.
\newblock Multilingual multi-aspect explainability analyses on machine reading comprehension models.
\newblock \emph{iScience}, 25\penalty0 (5):\penalty0 104176, 2022.
\newblock ISSN 2589-0042.
\newblock \doi{https://doi.org/10.1016/j.isci.2022.104176}.
\newblock URL \url{https://www.sciencedirect.com/science/article/pii/S2589004222004461}.

\bibitem[Cui et~al.(2023)Cui, Yang, and Yao]{chinese-llama-alpaca}
Yiming Cui, Ziqing Yang, and Xin Yao.
\newblock Efficient and effective text encoding for chinese llama and alpaca.
\newblock \emph{ArXiv preprint}, abs/2304.08177, 2023.
\newblock URL \url{https://arxiv.org/abs/2304.08177}.

\bibitem[Dettmers et~al.(2024)Dettmers, Pagnoni, Holtzman, and Zettlemoyer]{dettmers2024qlora}
Tim Dettmers, Artidoro Pagnoni, Ari Holtzman, and Luke Zettlemoyer.
\newblock Qlora: Efficient finetuning of quantized llms.
\newblock \emph{Advances in Neural Information Processing Systems}, 36, 2024.

\bibitem[Fedus et~al.(2022)Fedus, Zoph, and Shazeer]{fedus2022switch}
William Fedus, Barret Zoph, and Noam Shazeer.
\newblock Switch transformers: Scaling to trillion parameter models with simple and efficient sparsity.
\newblock \emph{The Journal of Machine Learning Research}, 23\penalty0 (1):\penalty0 5232--5270, 2022.

\bibitem[Hendrycks et~al.(2021)Hendrycks, Burns, Basart, Zou, Mazeika, Song, and Steinhardt]{hendrycks2020measuring}
Dan Hendrycks, Collin Burns, Steven Basart, Andy Zou, Mantas Mazeika, Dawn Song, and Jacob Steinhardt.
\newblock Measuring massive multitask language understanding.
\newblock In \emph{9th International Conference on Learning Representations, {ICLR} 2021, Virtual Event, Austria, May 3-7, 2021}. OpenReview.net, 2021.
\newblock URL \url{https://openreview.net/forum?id=d7KBjmI3GmQ}.

\bibitem[Huang et~al.(2023)Huang, Bai, Zhu, Zhang, Zhang, Su, Liu, Lv, Zhang, Lei, Fu, Sun, and He]{huang2023ceval}
Yuzhen Huang, Yuzhuo Bai, Zhihao Zhu, Junlei Zhang, Jinghan Zhang, Tangjun Su, Junteng Liu, Chuancheng Lv, Yikai Zhang, Jiayi Lei, Yao Fu, Maosong Sun, and Junxian He.
\newblock C-eval: A multi-level multi-discipline chinese evaluation suite for foundation models.
\newblock \emph{ArXiv preprint}, abs/2305.08322, 2023.
\newblock URL \url{https://arxiv.org/abs/2305.08322}.

\bibitem[Jiang et~al.(2024)Jiang, Sablayrolles, Roux, Mensch, Savary, Bamford, Chaplot, Casas, Hanna, Bressand, et~al.]{jiang2024mixtral}
Albert~Q Jiang, Alexandre Sablayrolles, Antoine Roux, Arthur Mensch, Blanche Savary, Chris Bamford, Devendra~Singh Chaplot, Diego de~las Casas, Emma~Bou Hanna, Florian Bressand, et~al.
\newblock Mixtral of experts.
\newblock \emph{ArXiv preprint}, abs/2401.04088, 2024.
\newblock URL \url{https://arxiv.org/abs/2401.04088}.

\bibitem[Lai et~al.(2017)Lai, Xie, Liu, Yang, and Hovy]{lai-etal-2017-race}
Guokun Lai, Qizhe Xie, Hanxiao Liu, Yiming Yang, and Eduard Hovy.
\newblock {RACE}: Large-scale {R}e{A}ding comprehension dataset from examinations.
\newblock In \emph{Proceedings of the 2017 Conference on Empirical Methods in Natural Language Processing}, pp.\  785--794, Copenhagen, Denmark, 2017. Association for Computational Linguistics.
\newblock \doi{10.18653/v1/D17-1082}.
\newblock URL \url{https://aclanthology.org/D17-1082}.

\bibitem[Li et~al.(2023)Li, Zhang, Koto, Yang, Zhao, Gong, Duan, and Baldwin]{li2023cmmlu}
Haonan Li, Yixuan Zhang, Fajri Koto, Yifei Yang, Hai Zhao, Yeyun Gong, Nan Duan, and Timothy Baldwin.
\newblock Cmmlu: Measuring massive multitask language understanding in chinese, 2023.

\bibitem[Lin et~al.(2022)Lin, Hilton, and Evans]{lin2021truthfulqa}
Stephanie Lin, Jacob Hilton, and Owain Evans.
\newblock {T}ruthful{QA}: Measuring how models mimic human falsehoods.
\newblock In \emph{Proceedings of the 60th Annual Meeting of the Association for Computational Linguistics (Volume 1: Long Papers)}, pp.\  3214--3252, Dublin, Ireland, 2022. Association for Computational Linguistics.
\newblock \doi{10.18653/v1/2022.acl-long.229}.
\newblock URL \url{https://aclanthology.org/2022.acl-long.229}.

\bibitem[Loshchilov \& Hutter(2019)Loshchilov and Hutter]{loshchilov2018decoupled}
Ilya Loshchilov and Frank Hutter.
\newblock Decoupled weight decay regularization.
\newblock In \emph{7th International Conference on Learning Representations, {ICLR} 2019, New Orleans, LA, USA, May 6-9, 2019}. OpenReview.net, 2019.
\newblock URL \url{https://openreview.net/forum?id=Bkg6RiCqY7}.

\bibitem[Open{AI}(2022)]{chatgpt}
Open{AI}.
\newblock Introducing chatgpt.
\newblock \url{https://openai.com/blog/chatgpt}, 2022.

\bibitem[{OpenAI}(2023)]{gpt-4}
{OpenAI}.
\newblock {GPT-4 Technical Report}.
\newblock \emph{ArXiv preprint}, abs/2303.08774, 2023.
\newblock URL \url{https://arxiv.org/abs/2303.08774}.

\bibitem[Peng et~al.(2023)Peng, Quesnelle, Fan, and Shippole]{peng2023yarn}
Bowen Peng, Jeffrey Quesnelle, Honglu Fan, and Enrico Shippole.
\newblock Yarn: Efficient context window extension of large language models.
\newblock \emph{ArXiv preprint}, abs/2309.00071, 2023.
\newblock URL \url{https://arxiv.org/abs/2309.00071}.

\bibitem[Rafailov et~al.(2024)Rafailov, Sharma, Mitchell, Manning, Ermon, and Finn]{rafailov2024direct}
Rafael Rafailov, Archit Sharma, Eric Mitchell, Christopher~D Manning, Stefano Ermon, and Chelsea Finn.
\newblock Direct preference optimization: Your language model is secretly a reward model.
\newblock \emph{Advances in Neural Information Processing Systems}, 36, 2024.

\bibitem[Rasley et~al.(2020)Rasley, Rajbhandari, Ruwase, and He]{deepspeed}
Jeff Rasley, Samyam Rajbhandari, Olatunji Ruwase, and Yuxiong He.
\newblock Deepspeed: System optimizations enable training deep learning models with over 100 billion parameters.
\newblock In Rajesh Gupta, Yan Liu, Jiliang Tang, and B.~Aditya Prakash (eds.), \emph{{KDD} '20: The 26th {ACM} {SIGKDD} Conference on Knowledge Discovery and Data Mining, Virtual Event, CA, USA, August 23-27, 2020}, pp.\  3505--3506. {ACM}, 2020.
\newblock URL \url{https://dl.acm.org/doi/10.1145/3394486.3406703}.

\bibitem[Sakaguchi et~al.(2020)Sakaguchi, Bras, Bhagavatula, and Choi]{sakaguchi2021winogrande}
Keisuke Sakaguchi, Ronan~Le Bras, Chandra Bhagavatula, and Yejin Choi.
\newblock Winogrande: An adversarial winograd schema challenge at scale.
\newblock In \emph{The Thirty-Fourth {AAAI} Conference on Artificial Intelligence, {AAAI} 2020, The Thirty-Second Innovative Applications of Artificial Intelligence Conference, {IAAI} 2020, The Tenth {AAAI} Symposium on Educational Advances in Artificial Intelligence, {EAAI} 2020, New York, NY, USA, February 7-12, 2020}, pp.\  8732--8740. {AAAI} Press, 2020.
\newblock URL \url{https://aaai.org/ojs/index.php/AAAI/article/view/6399}.

\bibitem[Shazeer(2020)]{swiglu}
Noam Shazeer.
\newblock Glu variants improve transformer, 2020.

\bibitem[Taori et~al.(2023)Taori, Gulrajani, Zhang, Dubois, Li, Guestrin, Liang, and Hashimoto]{alpaca}
Rohan Taori, Ishaan Gulrajani, Tianyi Zhang, Yann Dubois, Xuechen Li, Carlos Guestrin, Percy Liang, and Tatsunori~B. Hashimoto.
\newblock Stanford alpaca: An instruction-following llama model.
\newblock \url{https://github.com/tatsu-lab/stanford_alpaca}, 2023.

\bibitem[{Touvron} et~al.(2023){Touvron}, {Lavril}, {Izacard}, {Martinet}, {Lachaux}, {Lacroix}, {Rozi{\`e}re}, {Goyal}, {Hambro}, {Azhar}, {Rodriguez}, {Joulin}, {Grave}, and {Lample}]{llama}
Hugo {Touvron}, Thibaut {Lavril}, Gautier {Izacard}, Xavier {Martinet}, Marie-Anne {Lachaux}, Timoth{\'e}e {Lacroix}, Baptiste {Rozi{\`e}re}, Naman {Goyal}, Eric {Hambro}, Faisal {Azhar}, Aurelien {Rodriguez}, Armand {Joulin}, Edouard {Grave}, and Guillaume {Lample}.
\newblock Llama: Open and efficient foundation language models.
\newblock \emph{ArXiv preprint}, abs/2302.13971, 2023.
\newblock URL \url{https://arxiv.org/abs/2302.13971}.

\bibitem[Touvron et~al.(2023)Touvron, Martin, Stone, Albert, Almahairi, Babaei, Bashlykov, Batra, Bhargava, Bhosale, et~al.]{touvron2023llama2}
Hugo Touvron, Louis Martin, Kevin Stone, Peter Albert, Amjad Almahairi, Yasmine Babaei, Nikolay Bashlykov, Soumya Batra, Prajjwal Bhargava, Shruti Bhosale, et~al.
\newblock Llama 2: Open foundation and fine-tuned chat models.
\newblock \emph{ArXiv preprint}, abs/2307.09288, 2023.
\newblock URL \url{https://arxiv.org/abs/2307.09288}.

\bibitem[Vaswani et~al.(2017)Vaswani, Shazeer, Parmar, Uszkoreit, Jones, Gomez, Kaiser, and Polosukhin]{attention_is_all_you_need}
Ashish Vaswani, Noam Shazeer, Niki Parmar, Jakob Uszkoreit, Llion Jones, Aidan~N. Gomez, Lukasz Kaiser, and Illia Polosukhin.
\newblock Attention is all you need.
\newblock In Isabelle Guyon, Ulrike von Luxburg, Samy Bengio, Hanna~M. Wallach, Rob Fergus, S.~V.~N. Vishwanathan, and Roman Garnett (eds.), \emph{Advances in Neural Information Processing Systems 30: Annual Conference on Neural Information Processing Systems 2017, December 4-9, 2017, Long Beach, CA, {USA}}, pp.\  5998--6008, 2017.
\newblock URL \url{https://proceedings.neurips.cc/paper/2017/hash/3f5ee243547dee91fbd053c1c4a845aa-Abstract.html}.

\bibitem[Zellers et~al.(2019)Zellers, Holtzman, Bisk, Farhadi, and Choi]{zellers2019hellaswag}
Rowan Zellers, Ari Holtzman, Yonatan Bisk, Ali Farhadi, and Yejin Choi.
\newblock {H}ella{S}wag: Can a machine really finish your sentence?
\newblock In \emph{Proceedings of the 57th Annual Meeting of the Association for Computational Linguistics}, pp.\  4791--4800, Florence, Italy, 2019. Association for Computational Linguistics.
\newblock \doi{10.18653/v1/P19-1472}.
\newblock URL \url{https://aclanthology.org/P19-1472}.

\end{thebibliography}
\bibliographystyle{iclr2023_conference}


\end{CJK*}
\end{document}